%% file: 0-main.tex
\documentclass[10pt,twocolumn,letterpaper]{article}

\usepackage{iccv}
\usepackage{times}
\usepackage{epsfig}
\usepackage{graphicx}
\usepackage{amsmath}
\usepackage{amssymb}
\usepackage{eucal}

\usepackage{booktabs}
\usepackage{balance}
\usepackage[T1]{fontenc}

\usepackage{url}
\usepackage{bbm}
\usepackage{tabularx}
\usepackage{algorithm}
\usepackage[noend]{algpseudocode} 
\usepackage{enumitem}
\usepackage{caption}
\usepackage{xspace}
\usepackage{xcolor}
\usepackage{subcaption}
\usepackage{wrapfig,lipsum,booktabs}
\usepackage{amssymb}
\usepackage{comment}
\usepackage{xcolor}
\usepackage{color, colortbl}

\usepackage[pagebackref=true,breaklinks=true,colorlinks,linkcolor=blue,citecolor=blue,bookmarks=false]{hyperref}

\iccvfinalcopy 

\def\iccvPaperID{4464} 

\definecolor{LightCyan}{rgb}{0.88,1,1}

\ificcvfinal\fi
\newcommand{\algname}{{Prompt-OVD}}

\begin{document}

\title{\!\!Prompt-Guided Transformers for End-to-End Open-Vocabulary Object Detection\!\!}

\author{
Hwanjun Song~\thanks{Hwanjun and Jihwan contributed equally and this work was done while working at NAVER AI Lab.}~~\thanks{Hwanjun is also the corresponding author (hwanjuns@amazon.com).}\\
AWS AI Labs\\
{\tt\small hwanjuns@amazon.com}
\and 
Jihwan Bang~\footnotemark[1]\\
NAVER Cloud\\
{\tt\small jihwan.bang@navercorp.com}
}

\maketitle
\ificcvfinal\thispagestyle{empty}\fi

\begin{abstract}
Prompt-OVD is an efficient and effective framework for open-vocabulary object detection that utilizes class embeddings from CLIP as prompts, guiding the Transformer decoder to detect objects in both base and novel classes. Additionally, our novel RoI-based masked attention and RoI pruning techniques help leverage the zero-shot classification ability of the Vision Transformer-based CLIP, resulting in improved detection performance at minimal computational cost. Our experiments on the OV-COCO and OV-LVIS datasets demonstrate that Prompt-OVD achieves an impressive 21.2 times faster inference speed than the first end-to-end open-vocabulary detection method (OV-DETR), while also achieving higher APs than four two-stage-based methods operating within similar inference time ranges. Code will be made available soon. 
\end{abstract}

\input{1-introduction}
\input{2-relatedwork}

\input{3-method}
\input{4-experiment}

\input{5-conclusion}
{\small
\bibliographystyle{ieee_fullname}
\bibliography{reference}
}

\clearpage
\begin{appendix}
\input{6-appendix}
\end{appendix}

\end{document}

%% file: 1-introduction.tex
\section{Introduction}

Object detection is one of crucial tasks for real-world computer vision applications, which localizes and detects visible objects from the scenes\,\cite{he2017mask, padilla2020survey, papageorgiou1998general, redmon2016you}. Recently, the development of visual language models has enabled significant progress towards open-vocabulary object detection\,(OVD), which can detect novel classes that were \emph{not} visible during training\,\cite{guopen2022vild, zang2022open, zhong2022regionclip}. The large visual language models like CLIP\,\cite{radford2021clip} help recognize the concept of common objects using their image-text aligned embeddings. Therefore, it offers a more practical and effective detection pipeline by eliminating the need of re-training object detectors on newly collected data. 

\begin{figure*}[t!]
\begin{center}
\includegraphics[width=15cm]{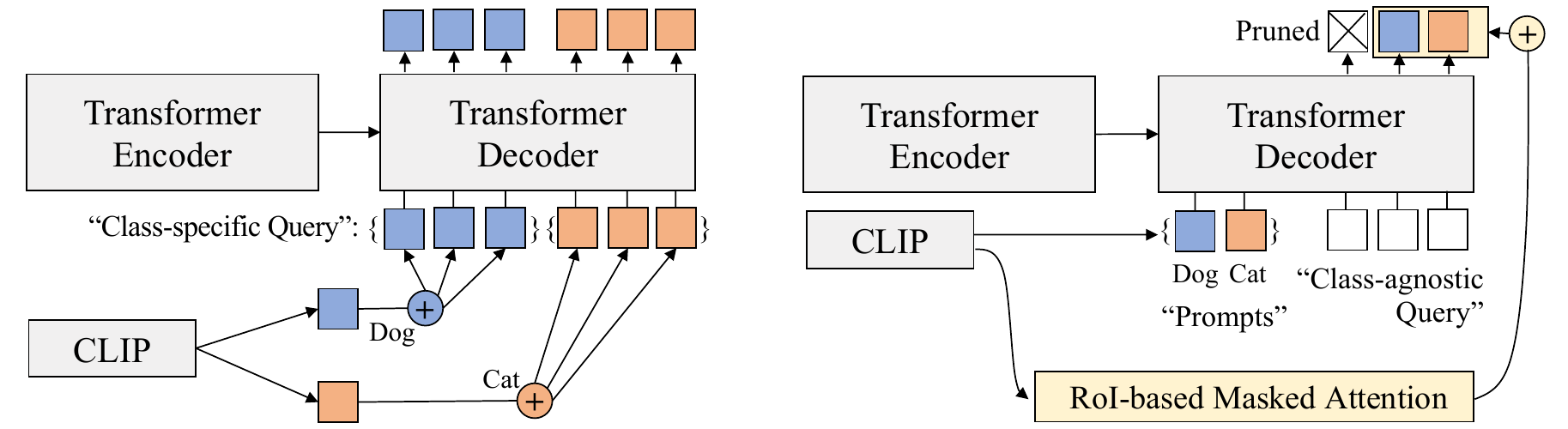}
\end{center}
\vspace*{-0.2cm}
\hspace*{4.0cm}{\small (a) OV-DETR.} \hspace*{5.2cm}{\small (b) Prompt-OVD (Ours).}
\vspace*{-0.1cm}
\caption{Architectural comparison: (a) OV-DETR needs linearly increasing number of object queries with respect to the number of classes and (b) \algname{} maintains a constant number of object queries by using class prompts.}
\label{fig:overview}
\vspace*{-0.4cm}
\end{figure*}

Most OVD approaches heavily rely on the region proposal network\,(RPN), which can easily overfit to base classes\,\cite{guopen2022vild, zhong2022regionclip}, and several approaches have extended it to leverage external data to mitigate this limitation\,\cite{minderer2022owlvit, rasheedbridging, zhou2022detic}. On the other hand, OV-DETR\,\cite{zang2022open} proposes a Transformer-based framework with conditional matching, which has two advantages over RPN-based methods.
First, it has the potential to localize novel objects better than the RPN by inject \emph{class-specific} knowledge, i.e., the addition of CLIP embeddings to the object queries as in Figure \ref{fig:overview}(a), to the Transformer decoder. 
Second, it can be trained in an \emph{end-to-end} fashion, eliminating the need for handcraft components like anchor generation and non-maximum suppression\,\cite{carion2020end, zhu2020deformable}. However, the main limitation of OV-DETR is the need for multiple forward passes in the decoder due to its large number of object queries, which results in slow inference speeds that may not be suitable for real-world use cases.

In this paper, we propose an efficient yet effective end-to-end Transformer-based framework named \textbf{Prompt-OVD}, as shown in Figure \ref{fig:overview}(b). Our framework leverages class embeddings from CLIP as \emph{class prompts} to the Transformer decoder. Unlike OV-DETR, which forces all object queries to be class-specific by mathematically adding class embeddings, we keep the object queries class-agnostic by prepending the class embeddings as {class prompts} to the decoder. Hence, we can maintain the \emph{constant} number of object queries regardless of increasing number of classes, significantly expediting the inference speed when handling a large number of object classes like LVIS data. In contrast, as shown in Figure \ref{fig:overview}(a), OV-DETR needs linearly increasing number of object queries with respect to the number of classes.
The proposed class prompts provide class-specific instruction instantaneously to the decoder on which object classes should be detected. Therefore, the output object embeddings by the decoder can be generalized to classes given by the prompts, resulting in very high recall for box regression on base and novel object classes.

Furthermore, there is still a significant gap in classification accuracy between base and novel classes in OVD, as only the base classes are visible during the end-to-end training pipeline. To mitigate such gap, we benefit from the {zero-shot} classification ability of the Vision Transformer\,(ViT)-based CLIP, and propose two additional techniques to reduce its computational overhead: efficient \emph{RoI-based masked attention}, which extracts the CLIP embeddings of the given RoIs with a minimal cost of a single inference path, and \emph{RoI pruning}, which selects only the RoIs that have objects with high probability from the entire box predictions. We then ensemble the classification results from the ViT-based CLIP with those from our detection model.

We conduct comprehensive experiments on two popular OVD datasets, namely OV-COCO\,\cite{lin2014microsoft} and OV-LVIS\,\cite{gupta2019lvis}, and compare \algname{} with an end-to-end method OV-DETR (baseline), but also four RPN-based two-phase OVD methods. The results demonstrate that \algname{} has an inference speed \emph{21.2 times faster} than OV-DETR and similar to that of the existing two-stage methods. Additionally, \algname{} achieves 30.6 $\mathrm{AP^{50}}$ of novel classes for OV-COCO and 29.4 $\mathrm{AP^{box}}$ of novel classes for OV-LVIS. Our contributions are as follows:
\begin{itemize}
\item We propose a prompt-based decoding that can keep a constant number of object queries, reducing the computational overhead of the Transformer decoder.  
\vspace*{-0.1cm}
\item We propose RoI-based masked attention and RoI pruning to benefit from a pre-trained ViT-based CLIP at minimal computational cost.
\vspace*{-0.1cm}
\item We considerably improve the efficiency and accuracy of the end-to-end Transformer-based detection pipeline for open vocabulary object detection.
\end{itemize}



%% file: 2-relatedwork.tex
\section{Related Work}

\noindent\textbf{Object Detection with Transformers.} 
Transformers\,\cite{vaswani2017attention} have emerged as one of the leading architectures for computer vision. The main advantage of using Transformers is its weaker inductive bias than convolutional neural networks\,(CNN), leading to the state-of-the-art results in image classification benchmarks\,\cite{dosovitskiy2020image, liu2021swin, han2022survey}. The use of Transformers has recently been extended to object detection and then rosolved the limitation of modern two-stage object detector like Mask-RCNN\,\cite{he2017mask} and YOLOS\,\cite{redmon2016you}. DETR\,\cite{carion2020end} eliminates the handcrafted components (e.g., anchor generation
and non-maximum suppression) by combining a CNN backbone and Transformer
encoder-decoders. Deformable DETR\,\cite{zhu2020deformable} introduces the notion of deformable attention to leverage multi-scale features as well as accelerating the slow training convergence of DETR. Next, YOLOS\,\cite{fang2021you} and ViDT\,\cite{songvidt} propose the fully Transformer-based object detection architecture without relying CNNs, proving the potential of ViTs as a generic object detector. In this paper, we present an end-to-end Transformer-based framework for OVD, distinct from the two-stage detectors.

\smallskip\smallskip
\noindent\textbf{Open Vocabulary Object Detection.} 
Despite significant progress in object detection~\cite{songvidt, fang2021you, he2017mask, redmon2016you}, training and scaling these models remains costly due to their closed-set assumption.
To address more object categories, including previously unseen ones, recent advances have leveraged visual language models, e.g., CLIP~\cite{radford2021clip}, to provide additional supervision for detection models. Their goal is open-vocabulary object detection\,(OVD), enabling the detection of previously unseen objects at the time of training.

One widely used approach in OVD is to employ a pre-trained CLIP model. ViLD~\cite{guopen2022vild} first proposes distilling knowledge from CLIP. Furthermore, DetPro~\cite{du2022detpro} suggests continuously learning prompts based on a pre-trained vision language model, and F-VLM~\cite{kuoopen2023fvlm} suggests that freezing CLIP weights as a backbone is more effective for novel object classes. This family of methods follows the pipeline of Mask-RCNN\,\cite{he2017mask}, inheriting the limitations of the two-stage detection model.

In contrast, OV-DETR~\cite{zang2022open} is the first to adopt DETR~\cite{carion2020end} for an end-to-end OVD. This involves formulating the learning objective as a binary matching problem between object queries and corresponding objects. However, OV-DETR requires significant computational power compared to Mask-RCNN or other two-stage detection methods. In addition, several approaches~\cite{zhong2022regionclip, rasheedbridging, zhou2022detic, minderer2022owlvit} have attempted to relax the problem to an unrestricted setup, which permits the use of external large-scale data that covers most common concepts in the real world, such as CC3M~\cite{sharma2018cc3m} and ImageNet-21k~\cite{deng2009imagenet}. However, comparing these methods with those under the restricted setup that does not allow the use of external data is unfair. It should be noted that our detection framework follows the restricted setup, requiring a stronger zero-shot generalization capability.



\section{Preliminary}
We briefly overview the pipelines of Transformer-based object detector and ViT-based CLIP.

\smallskip\smallskip
\noindent\textbf{Transformer-based Detector.} Detection Transformers (DETR) have replaced the RPN-based two-stage pipeline with a simple Transformer encoder-decoder pipeline\,\cite{carion2020end, zhu2020deformable, li2022exploring}, and its extensions have realized a fully Transformer-based object detector that removes the need for CNN backbones by adopting the pre-trained ViT as its Transformer encoder\,\cite{fang2021you, litransformer, songvidt}. For the fully Transformer architecture, the $L$-layer Transformer encoder receives an image as the input of {patch tokens}, $\mathrm{{I}_{0}=[patch_1, \dots, patch_{n}] \in {\mathbb{R}^{n \times d}}}$, which are progressively calibrated across the layers via self-attention, 
\begin{equation}
\begin{gathered}
{\rm I}_{L}={\rm Encoder}({\rm I}_{0}) \in \mathbb{R}^{n \times d}.
\end{gathered}
\end{equation} 
Subsequently, the Transformer decoder receives \emph{object queries} ${\rm Q}_{0}=[{\rm qeury}_1, \dots, {\rm qeury}_{m}]\in {\mathbb{R}^{m \times d}}$ that aggregate the key contents from the encoder output $\mathrm{I}_{L}$ to produce $m$ different object embeddings via alternating self-attention and cross-attention, 
\begin{equation}
\mathrm{{O}}_{L}={\rm Decoder}({\rm Q}_{0}, {\rm I}_{L})\in \mathbb{R}^{m \times d}.
\label{eq:dec_output}
\end{equation}
Lastly, the object embeddings are directly fed to a 3-layer feedforward neural networks\,(FFNs) for bounding box regression and linear projection for classification, 
\begin{equation}
\begin{gathered}
\mathrm{\hat{B}_{det} =FFN_{3-layer}}({\rm O}_{L}) \in \mathbb{R}^{m \times 4}\\ \mathrm{\hat{P}_{det}=Linear}({\rm O}_{L})\in\mathbb{R}^{m \times k},
\end{gathered}
\label{eq:prediction}
\end{equation}
where $k$ is the number of object classes. The bipartite Hungarian matching is used to train the model in an end-to-end fashion \,\cite{carion2020end}.

\smallskip\smallskip
\noindent\textbf{ViT-based CLIP.} CLIP\,\cite{radford2021clip} is a dual-encoder architecture that consists of visual and text encoders, returning the image-text aligned embeddings of the visual and text inputs. ViT-based CLIP uses a ViT architecture as the visual encoder and a masked self-attention Transformer as the text encoder. It receives the patch tokens of an image as input to the visual encoder, while image-aligned word tokens to the text encoder. Given an image ${\bf x}$ and a text prompt  ${\bf y^{cls}}$ for its class name, CLIP generates the two output embeddings, an image embedding ${\bf e}^{\rm img}_{\rm clip}\in \mathbb{R}^{d^{\prime}}$ and a text embedding ${\bf e}^{\rm txt}_{\rm clip}\in \mathbb{R}^{d^{\prime}}$ as:
\begin{equation}
{\bf e}^{\rm img}_{\rm clip} = {\rm CLIP}_{\rm img}({\bf x}) ~~{\rm and}~~ {\bf e}^{\rm txt}_{\rm clip} = {\rm CLIP}_{\rm txt}({\bf y^{cls}}).
\end{equation}
Since the two output embeddings are aligned each other via contrastive learning, knowledge distillation using CLIP embeddings has been shown to improve the detection performance on previsouly unseen classes\,\cite{guopen2022vild, zang2022open}.

%% file: 3-method.tex
\section{\algname{}}

We first reconfigure the decoding procedure of DETR to leverage the CLIP embeddings as its class prompts. Next, we propose an efficient way of ensembling with CLIP using RoI-based masked attention and RoI pruning.  

\subsection{Prompt-based Decoding}

Our first objective is designing a highly efficient decoding pipeline for an end-to-end Transformer-based open-vocabulary detector. To achieve this, we propose the concept of \emph{prompt-based decoding}. Similar to OV-DETR\,\cite{zang2022open}, our approach involves selecting either an image embedding ${\bf e^{img}}$ or a text embedding ${\bf e^{txt}}$ from CLIP to provide prior information about the classes that the decoder detects. However, we take this approach further by treating the two embeddings as \emph{class prompts}. These prompts modifies object queries ${\rm Q}$ in real-time via self-attention, enabling them to instantly detect class objects specified by the prompts.

\begin{figure}[t!]
\begin{center}
\includegraphics[width=8.6cm]{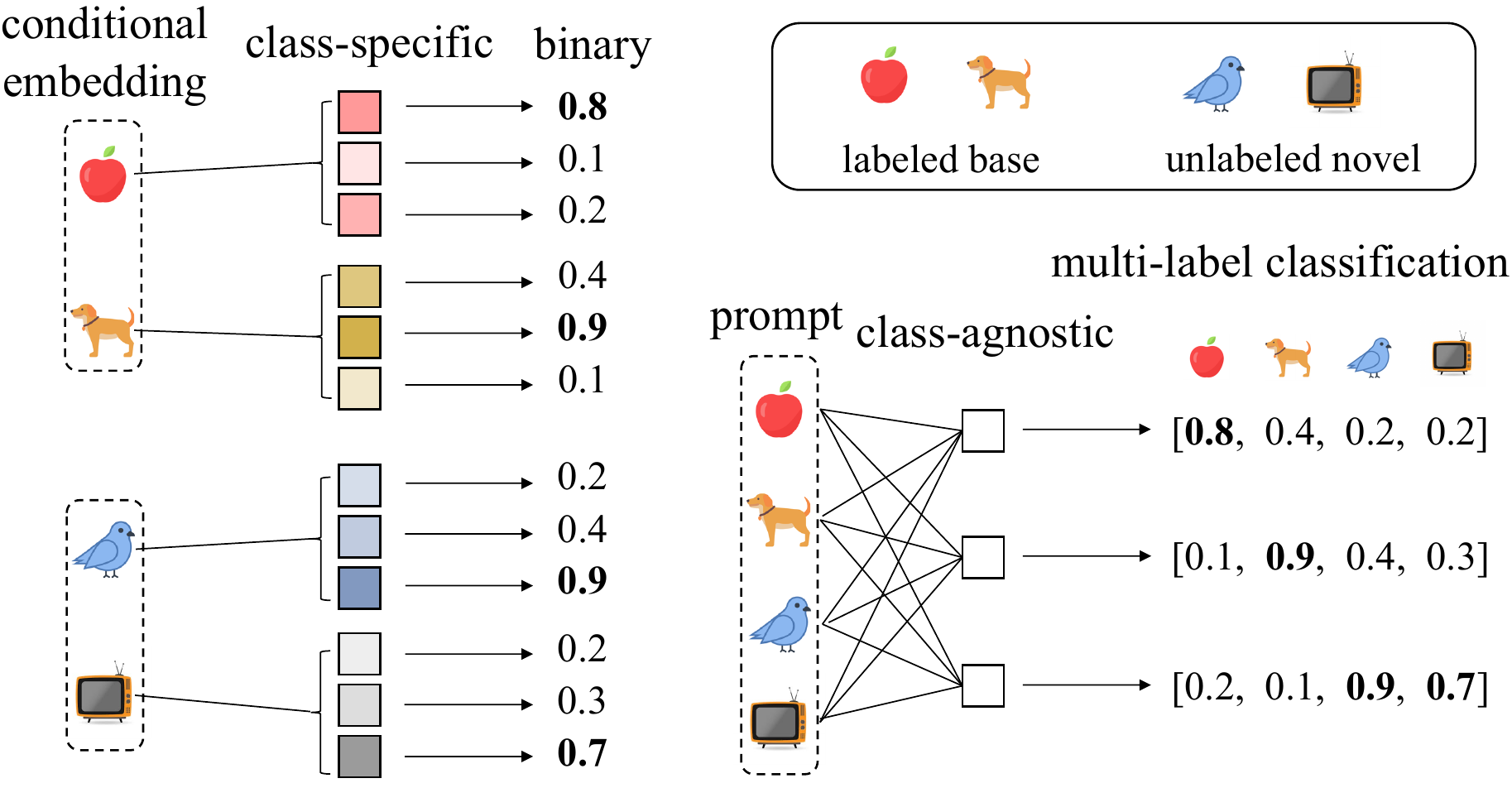}
\end{center}
\vspace*{-0.25cm}
\hspace*{1.0cm}{\small (a) OV-DETR.} \hspace*{2.3cm}{\small (b) Prompt-OVD.}
\vspace*{-0.15cm}
\caption{Difference in label assignment after decoding, where \# of classes $k$ is 4 and \# of object queries $m$ is 3. }
\label{fig:decoding}
\vspace*{-0.3cm}
\end{figure}

\smallskip\smallskip
\noindent\textbf{Decoding.} The decoder in DETR is a structure of alternating self-attention and cross-attention modules in each layer. Thus, we prepend the CLIP embedding as the prompts to the self-attention module of every decoding layer,
\begin{equation}
\begin{gathered}
{{\rm Q}_{l}^{\prime} = {\rm SelfAttn}_{l}([ {\rm \mathbb{F}}_{\rm proj}({\rm {\bf e}_{clip}^{mod}}) ; {\rm Q}_{l}])} \in \mathbb{R}^{m \times d}\\
{\rm mod} \in {\rm \{txt, img\}},
\label{eq:self_attn}
\end{gathered}
\end{equation}
where $\mathbb{F}_{\rm proj}$ is the projection layer to ensure that the CLIP embedding dimension $d^{\prime}$ matches the Transformer's embedding dimension $d$. In Eq.\,\eqref{eq:self_attn}, we prepend only a single clip embedding for simplicity, but multiple clip embeddings of different classes can be prepended in a batch (e.g., 10 prompts for 10 different classes). This results in the object queires ${\rm Q}^{\prime}$ becoming \emph{class-agnostic} for the classes given by prompts. They are then fed as input to the cross-attention module in the same layer, aggregating key contents from the encoder output ${\rm I}_{L}$ to represent objects,
\begin{equation}
\begin{gathered}
{{\rm O}_{l} = {\rm CrossAttn}_{l}({\rm Q}_{l}^{\prime}, {\rm I}_{L})} \in \mathbb{R}^{m \times d}.
\end{gathered}
\end{equation}
The object embeding ${\rm O}_{l}$ is also used as the object query ${\rm Q}_{l+1}$ to the next layer and, therefore, the final object embedding ${\rm O}_{L}$ is the output of the last decoding layer.


\smallskip\smallskip
\noindent\textbf{Prediction.} Figure \ref{fig:decoding} contrasts the difference in label assignment after decoding of OV-DETR and \algname{}. Specifically, OV-DETR forces all the object queries to be class-specific for {binary} classification, which requires input query tokens\,(i.e., the input of Transformer decoding layers) whose number is linearly proportional to the number of detected classes. For example, when using LVIS data with 1,203 classes, it requires $1,203 \times m$ object embeddings where $m = 1,500$ by default. On the other hand, \algname{} uses class-agnostic queries with prompts, which turns binary classification into a \emph{multi-label} classification.\footnote{Multi-label classification is a generalization of multi-class classification. Here, the labels are nonexclusive and there is no constraint on how many of the classes each RoI (bounding box) can include.} This requires only $m$ object embeddings but having identical number of predictions, i.e., $k \times m$. As for the box regression, each object embedding produces a single bounding box.

\subsection{Ensemble with ViT-based CLIP}

\begin{figure}[t!]
\begin{center}
\includegraphics[width=8.6cm]{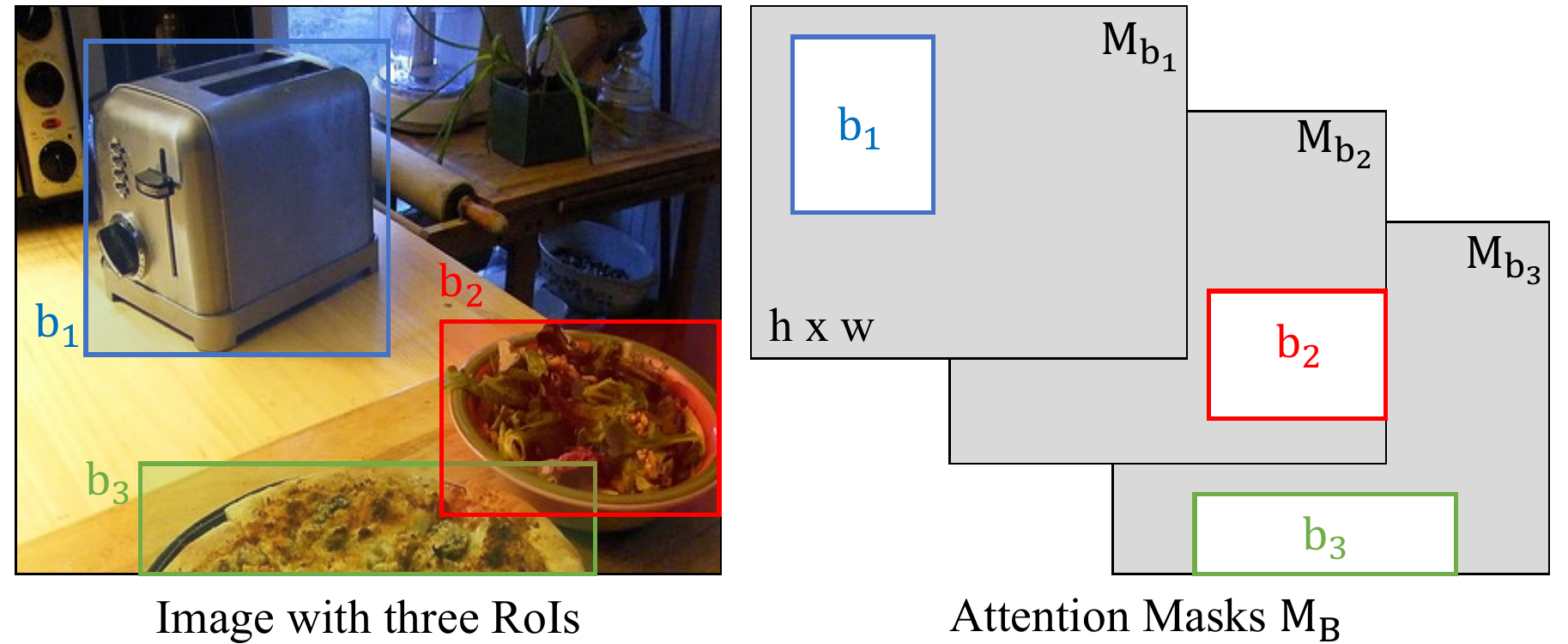}
\end{center}
\vspace*{-0.5cm}
\caption{Attention mask generation with RoIs. }
\label{fig:masks}
\vspace*{-0.4cm}
\end{figure}

Our second objective is to improve the classification performance on base and novel classes by utilizing the \emph{zero-shot} classification ability of the ViT-based CLIP. Let ${\rm \hat{B}_{det}}=\{{\bf b}_1, \dots {\bf b}_{m}\}$ denote the set of $m$ bounding boxes predicted by the prompt-based decoding as in Eq.\,\eqref{eq:prediction}. For zero-shot classification, a naive approach is repeatedly inferring on the cropped image for all RoIs as: 
\begin{equation}
\forall_{1 \leq i \leq m} {\bf e}^{\rm img}_{i} = {\rm CLIP}_{\rm img}\big({\rm Crop}({\bf x}, {\bf b}_i)\big)\in \mathbb{R}^{d^{\prime}},
\label{eq:iter_infer}
\end{equation}
and calculation of cosine similarity between the embedding ${\bf e}^{\rm img}_{i}$ of the $i$-th cropped image and text class embeddings ${\bf e}^{\rm txt}_{y}$, returning class prediction probabilities as:
\begin{equation}
\begin{gathered}
{\rm \hat{P}}_{\rm clip} = [\hat{\bf p}_{1}, \dots, \hat{\bf p}_{m}] \in \mathbb{R}^{m \times k}\\
\!\!\!\hat{\bf p}_{i}\! = \!{\rm Softmax}\big( \frac{1}{\tau} [ {\rm cos}({\bf e}^{\rm img}_{i}, {\bf e}^{\rm txt}_{1}), \dots, {\rm cos}({\bf e}^{\rm img}_{i}, {\bf e}^{\rm txt}_{k}) ] \big), \!\!\!
\end{gathered}
\end{equation}
where $\tau$ is the temperature on the logits.

Inspired by \cite{kuoopen2023fvlm}, we combine the CLIP class probabilities with the detection class probabilities to improve overall classification performance. The final class probabilities are given by: 
\begin{equation}
\!\!\!\!{\rm {P}}_{\rm final}{[\cdot, j]}\!= \!
\begin{cases}
\alpha {\rm {P}}_{\rm det}{[\cdot, j]} + (1\!-\!\alpha) {\rm {P}}_{\rm clip}{[\cdot, j]} &\!\!\!\! \text{if $j \in C_{B}$}\!\!\!\!\!\! \\
\beta {\rm {P}}_{\rm det}{[\cdot, j]} + (1\!-\!\beta) {\rm {P}}_{\rm clip}{[\cdot, j]}  &\!\!\!\! \text{if $j \in C_{N}$},\!\!\!\!\!\!
\end{cases}
\label{eq:ensemble}
\end{equation}
where $[\cdot, j]$ is the index for the $j$-th class column in the probability matrix ${\rm P} \in \mathbb{R}^{m \times k}$; $C_B$ and $C_N$ are the set of base and novel classes; and $\alpha, \beta \in [0, 1]$ control the probability weights for base and novel classes, respectively. This ensemble approach helps mitigate the classification performance gap between base and novel classes in open-vocabulary detection because novel classes benefit much more than base classes. The analysis can be found demonstrated in Section \ref{sec:abl_study}.


\subsubsection{RoI-based Masked Attention}

The iterative inference of Eq.\,\eqref{eq:iter_infer} harms the efficiency of using the large ViT-based CLIP model. RoI-Align \cite{he2017mask} is an efficient operator for extracting the feature maps from all the cropped patches, but this operator is confined to CNN architectures. Therefore, we introduce a new concept of \emph{RoI-based masked attention}, extracting the feature maps with a minimal cost of a single inference step.  

Let ${\rm W}_{Q}, {\rm W}_{K}, {\rm W}_{V}$ be the query, key, value projection matrices of the \emph{last} attention layer in the ViT. Then, the attention map ${\rm Attn}({\rm E}_Q, {\rm E}_K)\in \mathbb{R}^{hw}$ of the last Transformer layers is computed by:
\begin{equation}
\begin{gathered}
\!\!\!{\rm Attn}=  {\rm Softmax} \big(\frac{({\rm E}_Q {\rm W}_{Q}) ({\rm E}_K{\rm W}_{K})^{\top}}{\sqrt{d}} \big) \!\!\!\\
{\rm where} ~~ {\rm E}_Q = [{\bf e}^{\rm cls}_{L-1}] ~~ {\rm and} ~~ {\rm E}_K = [{\bf e}^{\rm cls}_{L-1}; \rm{I}_{L-1}].
\end{gathered}
\label{eq:pruning}
\end{equation}
${\bf e}^{\rm cls}_{L-1}$ is the class token and ${\rm I}_{L-1}$ is the patch tokens from the ($L$-1)-th Transformer layer. Note that the query ${\rm E}_Q$ includes only the class token since the clip image embedding is the projection of the class token's output\,\cite{radford2021clip}. We convert this attention process to the {RoI-based masked attention}. As seen in Figure \ref{fig:masks}, we generate attention masks ${\rm M}_{\hat{\rm B}} = [{\rm M}_{{\bf b}_{1}}, \dots, {\rm M}_{{\bf b}_{m}}] \in \mathbb{R}^{m \times hw}$; ${\rm M}_{{\bf b}_{i}}$ has the $0$ values for the area that overlaps the given RoI ${\bf b}_{i}$, -100 otherwise. Each mask penalizes the patch tokens outside the corresponding RoI not to affect the attention operation as:
\begin{equation}
{\rm Softmax} \big(\frac{({\rm E}_Q {\rm W}_{Q}) ({\rm E}_K{\rm W}_{K})^{\top} \!\!+ {\rm M}_{{\bf b}_{i}}}{\sqrt{d}}  \big).
\end{equation}
This eliminate the need of iterative inference over all RoIs by simply applying the proposed RoI-based masked attention in the last Transformer layer. Thus, the inference on the image is now modified from Eq.\,\eqref{eq:iter_infer} to:
\begin{equation}
[{\bf e}^{\rm img}_{i},\dots,{\bf e}^{\rm img}_{m}] = {\rm CLIP}_{\rm img}({\bf x}, \hat{\rm B}_{\rm det})\in \mathbb{R}^{m \times d^{\prime}}.
\label{eq:new_clip_img}
\end{equation}

\subsubsection{RoI Pruning} 

To make better use of CLIP, it is necessary to provide only RoIs that likely contain target objects. As discussed in \cite{zhong2022regionclip}, the contrastive learning of CLIP is unaware of the alignment between local image regions and text tokens, leading to incorrect predictions for background regions. To address this, we propose a simple RoI pruning that identifies non-background RoIs from box predictions. We define an \emph{object score} as the maximum prediction probability values across all classes given by prompts. Given the box prediction ${\rm \hat{B}_{det}}$ and its class prediction ${\rm \hat{P}_{det}}$ from the Transformer decoder, we keep only the RoIs with an object score higher than a certain threshold $\epsilon$ as:
\begin{equation}
{\rm \hat{B}}_{\rm prune} = \{ {\bf b}_{i} \in {\rm \hat{B}}_{\rm det} : {\bf p}_{i} \in {\rm \hat{P}}_{\rm det} \wedge {\rm max} ({\bf p}_{i}) \geq \epsilon \}.
\end{equation}
Therefore, we replace the input of the RoI-based masked attention with the pruned RoIs.

\subsection{Training and Evaluation Pipeline}

For label assignment, \algname{} conducts multi-label classification using bipartite matching with the Hungarian algorithm\,\cite{lin2017focal}. Therefore, following the existing optimization pipeline in \cite{carion2020end, zhu2020deformable}, we uses focal loss $\ell_{\rm cls}$ for multi-label classification, and L1 loss $\ell_{\rm l1}$ and GIoU loss $\ell_{\rm iou}$\,\cite{rezatofighi2019generalized} for box regression. The embedding loss $\ell_{\rm embed}$ proposed by \cite{zang2022open} is also used to distil the CLIP's knowledge for open vocabulary object detection, 
\begin{equation}
\ell = \lambda_{\rm cls}\ell_{\rm cls} + \lambda_{\rm l1}\ell_{\rm l1} + \lambda_{\rm iou}\ell_{\rm iou} + \lambda_{\rm embed}\ell_{\rm embed}, 
\label{eq:naive_roi}
\end{equation}
where $\lambda$s are the balancing parameters. 

We set the number of object queries $m$ to 300 and 1,500 for MS-COCO~\cite{lin2014microsoft} and LVIS~\cite{gupta2019lvis}, respectively. Following the literature\,\cite{guopen2022vild, zang2022open, zhong2022regionclip, zhou2022detic, rasheedbridging}, only base classes are given in training time, while novel classes are given together with base classes in testing time. Appendix A provides further details on the training and evaluation pipeline.

%% file: 4-experiment.tex

\begin{table*}[t!]
\centering
\caption{{Main Results on OV-COCO and OV-LVIS:} We evaluate box AP with IoU threshold 0.5 ($\mathrm{mAP^{50}}$) on OV-COCO, and box AP ($\mathrm{mAP^{box}}$) and mask AP ($\mathrm{mAP^{mask}}$) on OV-LVIS. Note that $\mathrm{mAP_{novel}}$ and $\mathrm{mAP}$ indicate the performance of zero-shot and the entire of categories, respectively. Lastly, latency implies inference time per image in seconds for OV-LVIS.}
 \vspace*{-0.25cm}
\label{tab:main}
\resizebox{1.0\linewidth}{!}{%
\begin{tabular}{@{}lrrrrrrrrrrr@{}}
\toprule 
& & & & \multicolumn{3}{c}{OV-COCO}  & \multicolumn{5}{c}{OV-LVIS}  \\
\cmidrule(lr){5-7} \cmidrule(lr){8-12}  
{Methods}& Backbone & CLIP & Res. & $\mathrm{mAP^{50}_{novel}}$ & $\mathrm{mAP^{50}_{base}}$ & $\mathrm{mAP^{50}}$ & $\mathrm{mAP^{box}_{novel}}$ & $\mathrm{mAP^{box}}$ & $\mathrm{mAP^{mask}_{novel}}$ & $\mathrm{mAP^{mask}}$ & Latency\,(s)\\ 
\midrule
\multicolumn{1}{r}{\normalsize {\sffamily DETR-based}} \vspace*{0.1cm} \\
OV-DETR~\cite{zang2022open} & RN50 & ViT-B/32 & 1333  & 29.4 & 61.0 & 52.7 & 18.0 & 27.4 & 17.4 & 26.6 & 12.28 \\
\textbf{Prompt-OVD} & ViT-B/16 & ViT-L/14 & 840 & \textbf{30.6} & \textbf{63.5} & \textbf{54.9} & \textbf{29.4} & \textbf{33.0} & \textbf{23.1} & 24.2 & 0.58 \\
\cmidrule(lr){1-12}
\multicolumn{1}{r}{\normalsize {\sffamily RCNN-based}}&  \multicolumn{4}{r}{\normalsize \!\!\!\!\!\!\!\!{\sffamily(Latency Range: 0.40 -- 0.70 seconds)}}\vspace*{0.1cm} \\ 
Detic~\cite{zhou2022detic}     & RN50 & ViT-B/32 & 1333 & 27.8 & 51.0 & 45.0 & 23.6 & 30.4 & 21.4 & \textbf{26.9} & 0.47 \\
ViLD~\cite{guopen2022vild}      & RN50 & ViT-B/32 & 1333 & 27.6 & 59.6 & 51.3& 16.7 & 27.8 & 16.6 & 25.5 & 0.48  \\
F-VLM~\cite{kuoopen2023fvlm}     & RN50 & RN50 & 1024 & 28.0 & 43.7 & 39.6& 20.3 & 27.8 & 18.6 & 24.2 & 0.50 \\
DetPro~\cite{du2022detpro}    & RN50 & ViT-B/32 & 1333  & - & -& - & 20.8 & 28.4 & 19.8 & 25.9 & 0.67\\
\bottomrule
\end{tabular}%
}
\vspace*{-0.1cm}
\end{table*}

\begin{table*}[t!]

\parbox{0.3\linewidth}{
\centering
\caption{Latency change by modifying OV-DETR to \algname{}.}
\vspace*{-0.25cm}
\label{tab:inference_study}
\resizebox{1.0\linewidth}{!}{
\begin{tabular}{@{}llr@{}}
\toprule 
& {Modification}& Latency (s)\\ 
\midrule
 & OV-DETR & 12.28\\
\,\,\,(1)\!\! & ResNet $\xrightarrow{}$ ViT & 12.36\\ 
\,\,\,(2)\!\! & ViT $\xrightarrow{}$ ViTDet & 8.75\\ 
\,\,\,(3)\!\! & Prompt-based Decoding & 2.89\\
\,\,\,(4)\!\! & Ensemble with CLIP & 3.03\\
\,\,\,(5)\!\! & RoI Pruning ($\epsilon=0.3)$ & 0.58\\
\bottomrule
\label{table:modification}
\vspace*{-0.4cm}
\end{tabular}%
}}
{\color{white} \,}
\hfill
\parbox{0.33\linewidth}{
\centering
\caption{Performance with varying $\alpha$ when fixing $\beta=0.4$.}
\vspace*{-0.3cm}
\label{tab:alpha}
\resizebox{1.0\linewidth}{!}{%
\begin{tabular}{@{}lrrrr@{}}
\toprule 
{$\alpha$}& $\mathrm{mAP^{box}_{novel}}$ & $\mathrm{mAP^{box}}$ & $\mathrm{mAP^{mask}_{novel}}$ & $\mathrm{mAP^{mask}}$ \\ 
\midrule
0.0 & 28.1 & 30.8 & 21.9 & 22.4 \\
0.1 & 28.7 & 32.3 & 22.6 & 23.6\\
\textbf{0.2} & \textbf{29.4} & \textbf{33.0} & \textbf{23.1} & \textbf{24.2}  \\
0.3 & 29.5 & 32.2 & 23.2 & 23.7 \\
0.4 & 29.7 & 30.6 & 23.3 & 22.5 \\
0.5 & 29.9 & 28.8 & 23.5 & 21.2\\
1.0 & 30.5 & 14.5 & 24.0 & 10.8 \\
\bottomrule
\label{table:alpha_search}
\vspace*{-0.4cm}
\end{tabular}%
}}
{\color{white} \,}
\hfill
\parbox{0.33\linewidth}{
\centering
\caption{Performance with varying $\beta$ when fixing $\alpha=0.2$.}
\vspace*{-0.3cm}
\label{tab:beta}
\resizebox{1.0\linewidth}{!}{
\begin{tabular}{@{}lrrrr@{}}
\toprule 
{$\beta$}& $\mathrm{mAP^{box}_{novel}}$ & $\mathrm{mAP^{box}}$ & $\mathrm{mAP^{mask}_{novel}}$ & $\mathrm{mAP^{mask}}$ \\ 
\midrule
0.0 & 15.9 & 30.0 & 12.1 & 21.7\\
0.1 & 22.8 & 31.4 & 17.9 & 22.9 \\
0.2 & 29.0 & 32.7 & 22.5 & 23.9 \\
0.3 & 29.3 & 33.0 & 23.0 & 24.1 \\
\textbf{0.4} & \textbf{29.3} & \textbf{33.0} & \textbf{23.1} & \textbf{24.2} \\
0.5 & 28.6 & 33.0 & 22.4 & 24.1 \\
1.0 & 19.6 & 31.5 & 15.3 & 22.9 \\
\bottomrule
\label{table:beta_search}
\vspace*{-0.4cm}
\end{tabular}%
}
}

\vspace*{-0.35cm}
\end{table*}

\section{Evaluation} 

\noindent\textbf{Datasets.} We evaluate our approach on two popularly used benchmark datasets, namely OV-COCO and OV-LVIS, each of which is modified from MS-COCO~\cite{lin2014microsoft} and LVIS\,(v1)~\cite{gupta2019lvis}; OV-COCO has 121K images with 64 classes while OV-LVIS has 100K images with 1,203 classes.
Following OV-DETR~\cite{zang2022open}, COCO is split into 17 novel classes and 48 base classes. LVIS is split into three categories: 337 novel classes, and 866 common or base classes based on the number of training images. Note that we refer to the two datasets as OV-COCO and OV-LVIS, respectively, and only base classes are used for training.

\smallskip\smallskip
\noindent\textbf{Algorithms.} We compare \algname{} with an end-to-end OVD detection model named OV-DETR\,\cite{zang2022open} (baseline) and four two-stage OVD models. However, it should be noted that the two-stage OVD models are based on Mask-RCNN or allow the use of external large-scale data, which makes a fair comparison with the end-to-end Transformer-based detectors difficult. Therefore, to ensure a fair comparison, we follow two criteria: (1) the results should be obtained by only using base categories in training, i.e., the restricted OVD setup and (2) the models' inference speed should be in the range of 0.4~--~0.7 seconds/image, which is similar to that of our proposed framework. 


\smallskip\smallskip
\noindent\textbf{Implementation.} The proposed \algname{} builds upon Deformable DETR\,\cite{zhu2020deformable}, similar to OV-DETR. However, we merge the independent ResNet backbone and Transformer encoder into a single ViT encoder using ViTDet~\cite{li2022exploring}. As a result, our architecture is a purely Transformer encoder-decoder structure, following recent fully Transformer detection pipeline\,\cite{litransformer, songvidt}. 

For training, we initialize the backbone weights with a plain ViT backbone that has been pre-trained as Masked Autoencoders on ImageNet-1K. The entire model is then trained end-to-end for 50 epochs with a batch size of 32, a weight decay of 1e-4, and an AdamW optimizer. We set the initial learning rate to 2e-4 while using an image size of 840$\times$840. We implement and test all algorithms using PyTorch on eight NVIDIA V100 GPUs. 

For inference, there are three hyperparameters: the weights $\alpha$ and $\beta$ for ensembling in Eq.\,\eqref{eq:ensemble}; the threshold $\epsilon$ for RoI pruning in Eq.\,\eqref{eq:pruning}. The former weights are set to be $(0.2, 0.35)$ and $(0.2, 0.4)$ for OV-COCO and OV-LVIS, while the latter pruning threshold is set to 0.125 and 0.3 for OV-COCO and OV-LVIS. As for the ensemble for classification, we leverage the CLIP model that uses ViT-L/14 as its image encoder with a image size of 336$\times$336, which adds very little computational overhead with our RoI-based masked attention and RoI pruning. The detailed analysis of the hyperparameters and additional overhead due to using CLIP is provided in Section \ref{sec:abl_study}. 

In addition, we need to incorporate the mask head into our model for the evaluation on OV-LVIS, as RPN-based two-stage methods have reported both box and mask APs. Following the recent literature\,\cite{dong2021solq, song2022extendable}, we extend our DETR-based detector using SOLQ\,\cite{dong2021solq}, which can perform a joint training of object detection and instance segmentation by simply adding a unified query representation module. The mask vector size is set to be 1,024 while keeping remaining hyperparameters to be the same as SOLQ.

\smallskip\smallskip
\noindent\textbf{Evaluation Metrics.} We evaluate the detection accuracy of our method following exactly the same metrics used in prior OVD studies\,\cite{guopen2022vild, zang2022open, zhong2022regionclip, minderer2022owlvit}. Specifically, for OV-COCO, we use $\mathrm{mAP^{50}}$ which is a measure of the box average precision\,(AP) with an IoU threshold of 0.5. On the other hand, for OV-LVIS, we use both box mAP ($\mathrm{mAP^{box}}$) and mask mAP ($\mathrm{mAP^{mask}}$) obtained by the joint learning of object detection and instance segmentation, respectively. 

Inference time is also a crucial metric for practical applications. To compare the efficiency of different models, we compute the inference time of all methods using the same hardware environment, consisting of a single NVIDIA V100 GPU and six Intel(R) Xeon(R) Gold 5120 CPUs. To ensure an accurate inference time measurement, we calculate the average time of 100 iterations after five initial iterations, using a batch size of 1.

\subsection{Main Experiment}

We present a comprehensive comparison of \algname{} with other five OVD methods in terms of detection accuracy and speed. To ensure a fair comparison, we only include the results of RCNN-based methods that can operate at a similar inference speed as ours. Table \ref{tab:main} summarizes the results of \algname{} and other five OVD methods.

In general, \algname{} outperforms the previous end-to-end OVD method, OV-DETR, on both datasets. Notably, \algname{}'s inference speed is {$21.2$ times} faster than OV-DETR, thanks to its prompt-based decoding approach. Refer to Section \ref{sec:inference_speed} for an in-depth comparison of efficiency with OV-DETR. %
Moreover, Prompt-OVD exhibits superior performance in terms of box mAP, even compared to RCNN-based OVD methods. These results support the effectiveness of our design that utilizes ViT-based CLIP with RoI-based mask attention and RoI pruning, improving the overall performance. Further investigation of the two techniques can be found in Section~\ref{sec:masked_attention} and ~\ref{sec:pruning}.

Although we observe a larger gap between box mAP and mask mAP (29.4 $\mathrm{mAP^{box}_{novel}} \rightarrow 23.1 \mathrm{mAP^{mask}_{novel}})$, this is a result of inheriting the limitation of SOLQ\,\cite{dong2021solq}. Specifically, the vector encoding of 2D segmentation masks using discrete cosine transformation loses object details compared to the conventional FPN-style mask head\,\cite{song2022extendable}.  
Furthermore, despite using a larger ViT-B/16 backbone than ResNet-50, \algname{} exhibits comparable inference time to RCNN-based methods, thanks to its simple encoding-decoding pipeline. Therefore, our results demonstrate that \algname{} shows a potential of the end-to-end Transformer-based framework for OVD. 

In Appendices B and C, we discuss potential enhancements to our method and present the results of our experiments on using image queries other than text queries for open-vocabulary object detection, respectively.

\begin{figure*}[t]
\begin{center}
\includegraphics[width=16.7cm]{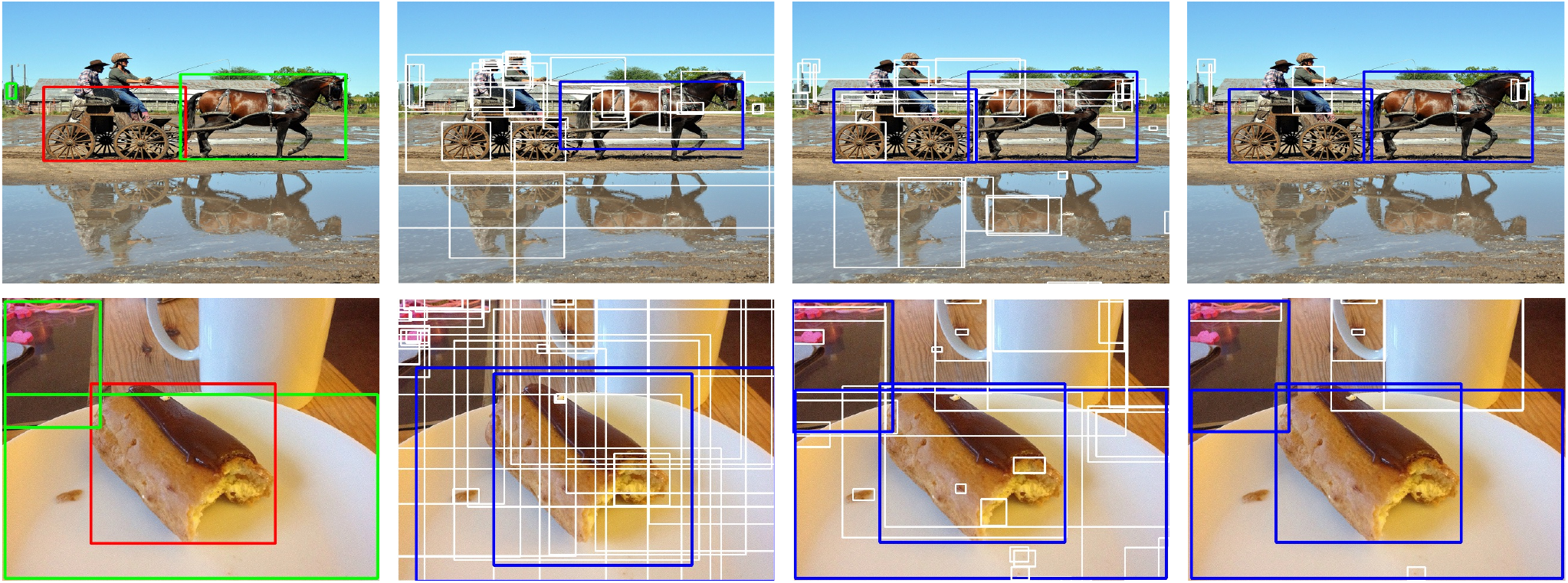}
\end{center}
\vspace*{-0.5cm}
\begin{subfigure}{0.3\textwidth}
\label{fig:gt}
\caption{GT}
\end{subfigure}
\begin{subfigure}{0.15\textwidth}
\label{fig:rpn}
\caption{RPN}
\end{subfigure}
\begin{subfigure}{0.31\textwidth}
\caption{No pruning ($\epsilon = 0$)}
\label{fig:noprune}
\end{subfigure}
\begin{subfigure}{0.16\textwidth}
\caption{Pruning ($\epsilon =0.3$)}
\label{fig:prune}
\end{subfigure}
\vspace*{-0.4em}
\caption{Box predictions: (a) ground-truth boxes, (b) boxes estimated by RPN from DetPro~\cite{du2022detpro}, (c)--(d) boxes estimated by \algname{} without and with RoI pruning. Due to numerous predicted boxes in (b) and (c), we limit the number of boxes to 40 for better visualization. Red and green boxes are the ground-truth of novel and base classes, while blue and white ones represent predicted boxes that are either in close proximity or not in close proximity to the ground-truth, respectively.}
\label{fig:pruning_abl}
\vspace*{-0.4cm}
\end{figure*}

\subsection{Main Ablation Study}
\label{sec:abl_study}

\subsubsection{Inference Speed Up from OV-DETR} 
\label{sec:inference_speed} 
Table \ref{table:modification} summarizes the change in inference speed when replacing each design component of OV-DETR with our proposed ones on OV-LVIS. (1) Despite having more parameters, using ViT-B/16 incurs little additional latency, as it rather reduces the computational burden for the multi-scale deformable attention of the DETR encoder. (2) The use of local attention and the replacement of the DETR encoder with a simple feature pyramid network, as suggested by \cite{li2022exploring, song2022extendable}, result in a meaningful reduction in latency. (3) The primary speedup comes from replacing the decoding of OV-DETR with prompt-based decoding. (4) The ensemble with ViT-based CLIP only adds very little latency thanks to our efficient RoI-based masked attention. (6) RoI pruning also significantly contributes to speeding up by reducing the number of RoI candidates for detection and segmentation, without sacrificing detection accuracy. Overall, \algname{} speeds up inference by $21.2$ times over OV-DETR.

\begin{table}[t!]
\centering

\caption{Performance between RoI Align and RoI-based Masked Attention (RMA) on OV-LVIS.}
\vspace*{-0.3cm}
\label{tab:roi_pool}
\resizebox{1.0\linewidth}{!}{%
\begin{tabular}{@{}lrrrrrr@{}}
\toprule 
{RoI Proc.}& $\mathrm{mAP^{box}_{novel}}$ & $\mathrm{mAP^{box}}$ & $\mathrm{mAP^{mask}_{novel}}$ & $\mathrm{mAP^{mask}}$ & Latency\\ 
\midrule
Naive & 12.8 & 28.3 & 9.9 & 20.3 & 13.51 \\
\hspace{3mm}+Pruning & 13.7 & 28.3 & 10.2 & 20.4 & 1.85\\
Align~\cite{he2017mask} & 24.2 & 31.7 & 19.1 & 23.0 & 3.06\\
\hspace{3mm}+Pruning & 26.2 & 32.0 & 20.8 & 23.4 & 0.60 \\
RMA (ours) & 26.6 & 32.5 & 21.0 & 23.8 & 3.03 \\
\hspace{3mm}+Pruning & \textbf{29.4} & \textbf{33.0} & \textbf{23.1} & \textbf{24.2} & 0.58 \\

\bottomrule
\end{tabular}%
}
\vspace*{-0.3cm}
\label{tab:rma_analysis}
\end{table}

\subsubsection{Ensemble Coefficient}
We investigate the influence of the ensembling weights $\alpha$ and $\beta$ for base and novel classes. We vary the value of each hyperparameter while keeping the other constant, as summarized in Tables \ref{tab:alpha} and \ref{tab:beta}, with a fixed RoI pruning threshold $\epsilon$ of 0.3. In general, the overall performance increases and then reaches at their maximum values when $\alpha=0.2$ and $\beta=0.4$. Using extreme values 0.0 or 1.0 for either coefficient results in significantly worse performance than using a more balanced ensemble. Moreover, the results show that the ensemble is more effective for novel classes than base classes, as evidenced by the significant impact of $\beta$ on the results of $\mathrm{mAP_{novel}}$. This suggests that the knowledge from CLIP has a greater positive impact on novel classes than on base classes. We set the values of $\alpha$ and $\beta$ to 0.2 and 0.4 for all experiments.
%


\subsubsection{RoI-based Masked Attention} 
\label{sec:masked_attention}
We conduct a comparison between our RoI-based masked attention method with both the naive approach and the commonly used RoI Align method, as summarized in Table \ref{tab:rma_analysis}. The naive approach implies that CLIP infers all the cropped images of RoIs according to Eq.~\eqref{eq:iter_infer}. To apply the RoI Align method to CLIP's ViT encoder, we reconstruct its patch tokens into a 2D feature map prior to the final Transformer layer. Compared to the naive approach, our RoI-based masked attention method has a significantly smaller computational overhead. Additionally, the efficiency and effectiveness of our method can be further improved by utilizing RoI pruning, which removes background RoIs. In contrast, RoI Align shows substantially lower $\mathrm{mAP^{box}_{novel}}$ and $\mathrm{mAP^{box}}$ than our method, as it is not optimized for the Transformer structure. Surprisingly, the naive crop method did not perform well, likely due to resizing a small object to be too large. Therefore, using masked attention is a more appropriate approach for Transformers than others. 

\label{sec:eval_RMA}
\begin{table}[t!]
\centering
\caption{Performance trade-off with varying $\epsilon$ on OV-LVIS.}
\vspace*{-0.2cm}
\label{tab:roi_thres}
\resizebox{1.0\linewidth}{!}{%
\begin{tabular}{@{}lrrrrr@{}}
\toprule 
{$\epsilon$}& $\mathrm{mAP^{box}_{novel}}$ & $\mathrm{mAP^{box}}$ & $\mathrm{mAP^{mask}_{novel}}$ & $\mathrm{mAP^{mask}}$ & Latency (s)\\ 
\midrule
0.0 & 26.6 & 32.5 & 21.0 & 23.8 & 3.03\\
0.1 & 27.7 & 32.8 & 21.9 & 24.0 & 1.38\\
0.2 & 28.3 & 33.0 & 22.3 & 24.1 & 0.86\\
\textbf{0.3} & \textbf{29.4} & \textbf{33.0} & \textbf{23.1} & \textbf{24.2} & 0.58\\
0.4 & 28.3 & 31.9 & 22.5 & 23.4 & 0.43\\
0.5 & 25.3 & 28.1 & 19.8 & 20.8 & 0.37\\
\bottomrule
\end{tabular}%
}
\vspace*{-0.2cm}
\end{table}

\subsubsection{Box Regression over RPN}
We validate the effectiveness of \algname{} in terms of box regression compared with the existing RPN-based method. Figure \ref{fig:pruning_abl} compares their estimated bounding boxes based on the ground-truth ones. 
Figure \ref{fig:pruning_abl}(b) is an example of the scenario where the RPN method fails to accurately localize objects, i.e., a  missing box for carriage (novel class) and two deviated boxes from the ground-truth for the bread (novel class) and plate (base class). In contrast, \algname{} successfully localizes all base and novel objects with high recall using the prompt-guided decoding, as shown in Figure \ref{fig:pruning_abl}(c). Despite the presence of background or inaccurate boxes in the box candidates, \algname{} successfully covers all the ground-truth boxes with its predictions. Furthermore, as seen in Figure \ref{fig:pruning_abl}(d), RoI pruning effectively excludes such irrelevant boxes from the detection process.

\subsubsection{RoI Pruning Threshold}
\label{sec:pruning}
We investigate the trade-off between detection performance and computational efficiency by varying the pruning threshold $\epsilon$, as summarized in Table \ref{tab:roi_thres}. As the threshold increases, fewer bounding boxes are retained, as the number of boxes with object scores greater than the threshold decreases. For instance, when the threshold is set to be 0.0, which means RoI pruning is not applied, the detection accuracy deteriorates due to the inclusion of background boxes, also resulting in a high latency. On the contrary, the detection accuracy improves as $\epsilon$ increases within the reasonable range of 0.0~--~0.3, but deteriorates with a larger threshold of 0.4~--~0.5. That is, as $\epsilon$ increases, more false positive boxes begins to be excluded, leading to improved performance. However, further increasing the threshold leads to the removal of true positive boxes, causing performance degradation. Therefore, we set the value of $\epsilon$ to 0.3 for all experiments.


\begin{table}[t!]
\centering
\caption{Performance when using different CLIP models for ensembling on OV-LVIS.}
\vspace*{-0.25cm}
\label{tab:clip_arch}
\resizebox{1.0\linewidth}{!}{%
\begin{tabular}{@{}lrrrrr@{}}
\toprule 
{CLIP}& $\mathrm{mAP^{box}_{novel}}$ & $\mathrm{mAP^{box}}$ & $\mathrm{mAP^{mask}_{novel}}$ & $\mathrm{mAP^{mask}}$ & Latency (s)\\ 
\midrule
None     & 15.4 & 28.1 & 11.7 & 20.3 & 0.54 \\
ViT-B/32 & 20.5 & 30.1 & 16.4 & 21.9 & 0.56 \\
ViT-B/16 & 22.3 & 31.1 & 17.5 & 22.6 & 0.57\\
\textbf{ViT-L/14} & \textbf{29.4} & \textbf{33.0} & \textbf{23.1} & \textbf{24.2} & 0.58 \\

\bottomrule
\end{tabular}%
}
\label{tab:clip_size}
\vspace{-1.2em}
\end{table}

\smallskip\smallskip
\subsection{Additional Design Choice}
\noindent We explore two supplementary design choices to utilize the ViT-based CLIP model in a manner that achieves the best balance between OVD performance and inference speed. 
We provide more supplementary analysis on applying RoI-based masked attention to different attention layers and using different pre-trained ViT backbones with \algname{} in Appendix D.

\subsubsection{CLIP Model Size} 
Table \ref{tab:clip_size} summarizes the performance obtained after the ensemble with three different sizes of ViT-based CLIP models, including the scenario where CLIP is not used at all. We observe that without using the ensemble technique, the zero-shot performance is very poor compared to when CLIP is employed. However, as the size of the CLIP model increases, both zero-shot and overall performance improve, albeit with slightly higher inference speed. The difference in inference time is negligible due to our proposed efficient techniques; RoI-based masked attention and RoI pruning. Therefore, we conclude that the benefits obtained from using a larger CLIP encoder with ViT-L/14 outweigh the minimal increase in inference time.


\subsubsection{CLIP Input Resolution} 
Another design consideration is the resolution of input to the ViT-based CLIP. To evaluate the performance trade-off between detection accuracy and latency of CLIP, we vary the input image size and report the results in Table \ref{tab:image_size}. Similar to the model size, the latency change by input resolution is negligible thanks to our efficient methodological design. This indicates that we can use a variety of image resolutions without sacrificing the latency of \algname{}. However, the best image resolution for the ViT-L/14 encoder is $336 \times 336$, which is the original input size used to train the CLIP model, while the detection accuracy for base and novel classes drops with a larger $672\times672$. The $336 \times 336$  provides a good balance between detection accuracy and inference time, so it is the recommended input resolution. 



\begin{table}[t!]
\centering
\caption{Performance when using different input image resolution for ViT-based CLIP on OV-LVIS.}
\vspace*{-0.25cm}
\label{tab:clip_img_res}
\resizebox{1.0\linewidth}{!}{%
\begin{tabular}{@{}lrrrrr@{}}
\toprule 
{img. res.}& $\mathrm{mAP^{box}_{novel}}$ & $\mathrm{mAP^{box}}$ & $\mathrm{mAP^{mask}_{novel}}$ & $\mathrm{mAP^{mask}}$ & Latency \\ 
\midrule

168$\times$168 & 28.5 & 30.1 & 20.9 & 22.0 & 0.57 \\ 
\textbf{336$\times$336} & \textbf{29.4} & \textbf{33.0} & \textbf{23.1} & \textbf{24.2} & 0.58 \\ 
672$\times$672 & 29.2 & 33.0 & 23.0 & 24.2 & 0.83 \\

\bottomrule
\end{tabular}%
}
\vspace*{-0.35cm}
\label{tab:image_size}
\end{table}




%% file: 5-conclusion.tex

\section{Conclusion} 
We introduce \algname{}, a novel object detection method that achieves highly efficient inference speed while also improving zero-shot generalization compared with existing methods. The prompt-based decoding approach reduces the computational burden of object queries. The RoI-based masked attention and RoI pruning techniques allow us to efficiently leverage a large ViT-based CLIP model, enhancing detection performance through classification prediction ensembling. Comprehensive experiments show that \algname{} is $21.2$ times faster than OV-DETR while achieving comparable or higher APs on base and novel classes compared to two-stage OVD methods. 


%% file: 6-appendix.tex
\section{Training and Evaluation Details}

\begin{figure*}[t]
\begin{center}
\includegraphics[width=16.7cm]{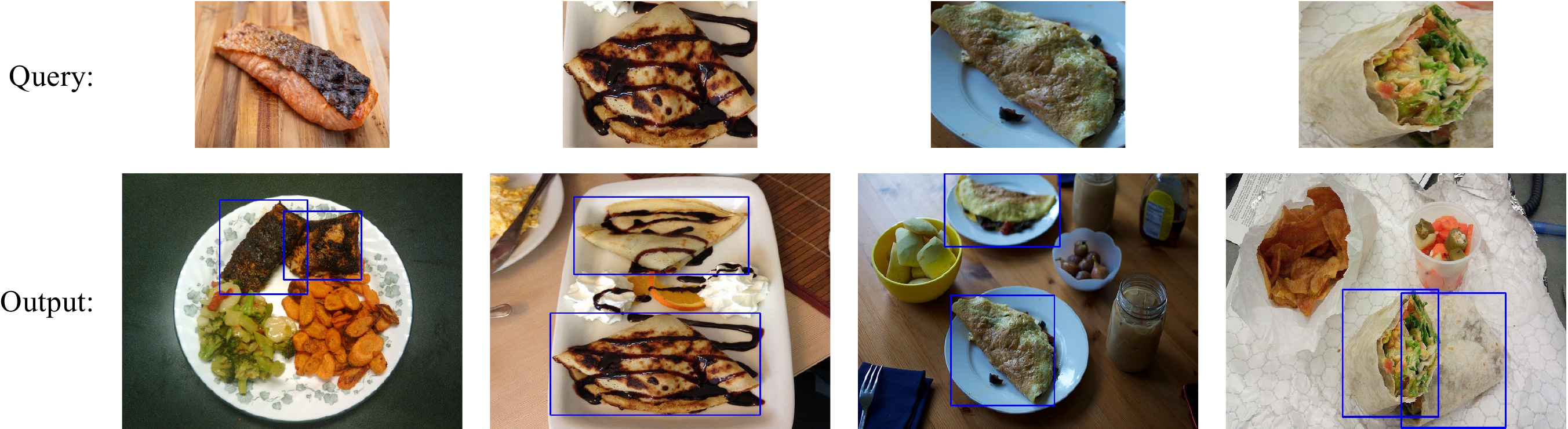}
\end{center}
\vspace*{-0.3cm}
\hspace{2cm} (a) Grilled salmon \hspace{1.8cm} (b) Crepe \hspace{2.2cm} (c) Omelette \hspace{2.1cm} (d) Burrito
\vspace*{-0.4em}
\caption{Image-conditioned open-vocabulary detection. Image queries are from novel classes and the sub-captions represent text corresponding to the image queries.}
\label{fig:image_conditioned}
\vspace*{-0.4cm}
\end{figure*}

We assess our method by conducting experiments on two widely used open-vocabulary object detection datasets, namely OV-COCO and OV-LVIS. Following the literature\,\cite{guopen2022vild, zang2022open}, we split the classes in MS-COCO into 48 base categories and 17 novel categories, where the remaining categories are not included since they do not belong to a synset in the WordNet hierarhiy\,\cite{zareian2021open}. Regarding OV-LVIS, we follow the setup of \cite{zareian2021open}, selecting 337 rare classes as novel categories and the remaining classes as common and frequent categories. Because of the difference in the number of object categories, OV-LVIS has a much larger number of objects to detect. Additionally, the number of object queries should be larger than the number of actual objects in an image. Therefore, the number of object queries is set to 300 and 1,500 for OV-COCO and OV-LVIS. 

\smallskip\smallskip
\noindent\textbf{Training Loss Function.}
\algname{} utilize the loss function from (Deformable) DETR\,\cite{carion2020end, zhu2020deformable}, which includes a detection head that generates a set of bounding boxes. We modify this for OVD such that the detection head only generates the boxes and class labels given by the class prompts. During the training phase, the number of class prompts can vary according to the number of visible base classes in the current mini-batch. 

Hungarian matching is employed to identify a bipartite matching between the predicted boxes ${\rm \hat{B}}$ and the ground-truth boxes ${\rm {B}}$. The training process basically involves three types of losses: a classification loss $\ell_{\rm cls}$, which is the focal loss\footnote{Multi-label classification is performed only for the classes given by the class prompts.}, a box distance loss $\ell_{\rm l1}$ and a generalized IoU loss $\ell_{\rm iou}$ for box regression as \,\cite{songvidt, zhu2020deformable}:
 \begin{equation}
\begin{gathered}
\ell_{\rm cls}(i) = -\text{log}~\hat{\rm {P}}_{\sigma(i),c_i},~~~\ell_{\rm  l1}(i) =  ||{\rm B}_{i}-\hat{\rm B}_{\sigma(i)}||_{1},\\
\ell_{\rm iou}(i) = 1 \!-\! \big( \frac{|{\rm B}_{i} \cap  \hat{{\rm B}}_{\sigma(i)}|}{|{\rm B}_{i} \cup  \hat{{\rm B}}_{\sigma(i)}|} - \frac{|{\sf {\rm B}}({\rm B}_{i}, \hat{{\rm B}}_{\sigma(i)}) \symbol{92} {\rm B}_{i} \cup  \hat{{\rm B}}_{\sigma(i)}| }{|{\sf {\rm B}}({\rm B}_{i}, \hat{{\rm B}}_{\sigma(i)})|} \big),
\end{gathered}
\end{equation}
where $c_i$ is the target base class and $\sigma(i)$ is the bipartite assignment of the $i$-th ground-truth box. Also, the embedding loss $\ell_{\rm embed}$ proposed by \cite{zang2022open} is used to distil the CLIP's knowledge for open vocabulary object detection. Therefore, the final objective of \algname{} is formulated as:
\begin{equation}
\ell = \lambda_{\rm cls}\ell_{\rm cls} + \lambda_{\rm l1}\ell_{\rm l1} + \lambda_{\rm iou}\ell_{\rm iou} + \lambda_{\rm embed}\ell_{\rm embed}, 
\label{eq:naive_roi}
\end{equation}
where $\lambda$s are the balancing parameters, where $\lambda_{\rm cls}=3, \lambda_{\rm l1}=5, \lambda_{\rm iou}=2$, and $\lambda_{\rm embed}=2$. 

\smallskip\smallskip
\noindent\textbf{Evaluation.}
In the testing phase, the class prompts are inclusive of all base and novel classes following the recent literature\,\cite{guopen2022vild, zang2022open, zhong2022regionclip, zhou2022detic, rasheedbridging}. For producing the final results, we apply RoI-based pruning to identify the RoIs with high object scores. Next, the classification results of the selected bounding boxes are ensembled with those from the ViT-based CLIP by Eq.\,\eqref{eq:ensemble}. 
\section{Potential Enhancement}

We discuss three possible ways to further enhance the potential of our framework. 

\smallskip\smallskip
\noindent\textbf{Prediction Ensemble.} In our ensemble method, we adopt an arithmetic mean approach as shown in Eq. \eqref{eq:ensemble}. However, this requires hyperparameter tuning to achieve the best performance, and the results are sensitive to the selection of $\alpha$ values, as indicated in Table \ref{tab:alpha}. The variation in performance highlights the importance of carefully choosing the hyperparameters for \algname{}, as suboptimal selections can have a significant impact on the overall performance. Although the simple ensemble approach provides a satisfactory detection accuracy for OV-COCO and OV-LVIS, we are of the belief that a more effective prediction ensemble strategy can be studied as future work to leverage both CLIP and a detection model in synergy.

\smallskip\smallskip
\noindent\textbf{Unrestricted Setup.} Our primary focus is on developing an end-to-end Transformer-based framework that is both efficient and effective. Therefore, we adopt a restricted setup in which external data is not permitted, as our aim is to investigate the potential of the framework itself in achieving optimal performance. Despite our restricted setup, we acknowledge that in real-world scenarios, large-scale external data are often available and their utilization can significantly enhance the zero-shot detection performance. We are confident that our framework can be extended in this direction, owing to its simple end-to-end encoding and decoding pipeline. However, we leave this as a potential avenue for future work, as our current focus is on investigating the framework's performance under the restricted setup.

\smallskip\smallskip
\noindent\textbf{Instance Segmentation.} A drawback of \algname{} is the considerable discrepancy in the performance between object detection and instance segmentation. While this is partly due to inheriting the limitations of using SOLQ\,\cite{dong2021solq} for end-to-end joint learning, the development of a more effective segmentation approach based on DETR will enhance the performance of our framework significantly. We leave this as an avenue for future research.

\section{Image Conditioned Detection}

One advantage of our framework is that it can detect objects using an {image query} instead of relying solely on the class name. This feature has the potential to identify the target object by using a cropped image of the object itself, even when we may not know the object's name. 

Figure \ref{fig:image_conditioned} shows the results of \algname{} when applying this image-conditioned object detection using four different image queries. Although their class names are not provided, \algname{} is capable of accurately localizing target objects that were not encountered during the training phase.  Additionally, \algname{} can recognize new classes even when the image query has a dissimilar shape of target objects, as depicted in Figure \ref{fig:image_conditioned}(a). These findings suggest that our algorithm is resilient in detecting objects using image queries for open-set scenarios.

\begin{table}[t!]
\centering
\caption{Performance when using different attention layers for RoI-based masked attention on OV-LVIS }
\vspace*{-0.25cm}
\label{tab:blk_num}
\resizebox{1.0\linewidth}{!}{%
\begin{tabular}{@{}lrrrr@{}}
\toprule 
{Attn. Layer Num.}& $\mathrm{mAP^{box}_{novel}}$ & $\mathrm{mAP^{box}}$ & $\mathrm{mAP^{mask}_{novel}}$ & $\mathrm{mAP^{mask}}$ \\
\midrule
$\mathrm{15^{th}}$ & 13.6 & 28.3 & 10.5 & 20.4 \\ 
$\mathrm{20^{th}}$ & 20.4 & 30.1 & 16.2 & 21.9 \\ 
\textbf{$\mathrm{24^{th}}$ (last layer)} & \textbf{29.4} & \textbf{33.0} & \textbf{23.1} & \textbf{24.2} \\
\bottomrule
\end{tabular}%
}
\label{tab:diff_layer}
\vspace*{-0.4cm}
\end{table}

\section{Supplementary Analysis}

We provide supplementary analysis on applying RoI-based masked attention to different attention layers in Appendix \ref{sec:diff_layer} and using different pre-trained ViT backbones with \algname{} in Appendix \ref{sec:diff_pretrain}.

\subsection{Pre-trained Weights.} 
\label{sec:diff_pretrain}

We analyze the impact of using different initial weights for the backbone, ImageNet and MAE, on the overall performance. As outlined in Table~\ref{tab:pretrained_model}, training the backbone using the MAE pre-trained weights yields better performance in dense prediction tasks such as object detection and instance segmentation, even under an open-vocabulary setup, compared to starting from the ImageNet pre-trained weights. This observation is consistent with previous research, such as \cite{li2022exploring}.
It is also worth noting that even when using the ImageNet pre-trained weights, \algname{} still outperforms the baseline\,(OV-DETR) with a much faster inference speed.
In conclusion, the choice of initial weight for the backbone plays a crucial role in the overall performance of the model, and using the MAE pre-trained weights as the starting point results in better performance.

\subsection{Different Layers for RoI-based Attention} 
\label{sec:diff_layer}

We investigate the impact of applying our RoI-based masked attention to different attention layers in the ViT-L/14 image encoder of CLIP. The results, as summarized in Table \ref{tab:diff_layer}, show that detection accuracy improves as the layer number increases, from the 15th to the 24th. Therefore, applying the technique to the last layer is the best design choice for open vocabulary detection with \algname{}.

\begin{table}[t!]
\centering
\caption{Performance when using different pre-trained ViT backbones on OV-LVIS.}
\vspace*{-0.25cm}
\label{tab:pretrained_model}
\resizebox{1.0\linewidth}{!}{%
\begin{tabular}{@{}lrrrr@{}}
\toprule 
{Pretrained Model}& $\mathrm{mAP^{box}_{novel}}$ & $\mathrm{mAP^{box}}$ & $\mathrm{mAP^{mask}_{novel}}$ & $\mathrm{mAP^{mask}}$ \\
\midrule
ImageNet~\cite{deng2009imagenet} & 26.4 & 28.7 & 20.7 & 20.9\\
\textbf{MAE}~\cite{he2022masked} & \textbf{29.4} & \textbf{33.0} & \textbf{23.1} & \textbf{24.2}\\
\bottomrule
\end{tabular}%
}
\end{table}